\documentclass{article}

\usepackage[preprint]{corl_2026} 

\usepackage{amsmath}
\usepackage{amssymb}
\usepackage{mathtools}
\usepackage{amsthm}
\usepackage{xspace}
\usepackage{multirow}
\usepackage{multicol}
\usepackage[table]{xcolor}

\usepackage{graphicx}
\usepackage{microtype}

\usepackage{booktabs}
\usepackage{hyperref}

\def\OURS{{GeoWorldAD}\xspace}

\title{GeoWorldAD: Geometry World Action Model for Autonomous Driving}

\author{
    Songyan Zhang$^{1,2}$, 
    Jinyuan Tian$^{3}$,
    Hanbing Li$^{2,\ddagger}$, Daqi Liu$^{2}$, Hao Chen$^{3}$, Wenhui Huang$^{1}$, \\[-0.1cm]
    \AND Fang Li$^{2}$, Guang Chen$^{2}$, Hangjun Ye$^{2}$, Long Chen$^{2}$, Kuiyuan Yang$^{2}$, Chen Lv$^{1,\dagger}$ \\[0.5cm]
    $^{1}$Nanyang Technological University,$^{2}$Xiaomi EV, $^{3}$Zhejiang University \\[-0.2cm]
}

\begin{document}
\maketitle

\renewcommand{\thefootnote}{\fnsymbol{footnote}}
\footnotetext[3]{Project Lead}
\footnotetext[2]{Corresponding Author}


\begin{abstract}

Autonomous driving requires both safe and efficient planning decisions in dynamic 3D environments. Although recent Vision/Video-Action models learn policies directly from visual observations and scale well with advances in vision transformers and large-scale training data, they often lack explicit geometric grounding and future-aware spatial guidance, limiting their ability to balance collision avoidance and driving progress. In this work, we propose GeoWorldAD, a geometry world action model that grounds trajectory planning in ego-aligned 3D space and anticipates short-horizon scene evolution with latent future geometry tokens. Present geometry provides essential spatial constraints for safe planning, while future geometry reveals how surrounding agents and ego-centric free space may evolve, reducing overly conservative decisions without sacrificing safety. To efficiently exploit these geometric cues, GeoWorldAD progressively aggregates multi-scale present geometry and latent future geometry through iterative trajectory refinement. Experiments on NAVSIM v1 and v2 demonstrate state-of-the-art performance, highlighting the effectiveness of explicit 3D geometry grounding and future geometry world modeling for safe and efficient autonomous driving.

\end{abstract}

\keywords{Autonomous Driving, Geometry Reconstruction, World Action Model} 


\section{Introduction}

\begin{figure*}[t]
\centering
\includegraphics[width=1.0\textwidth]{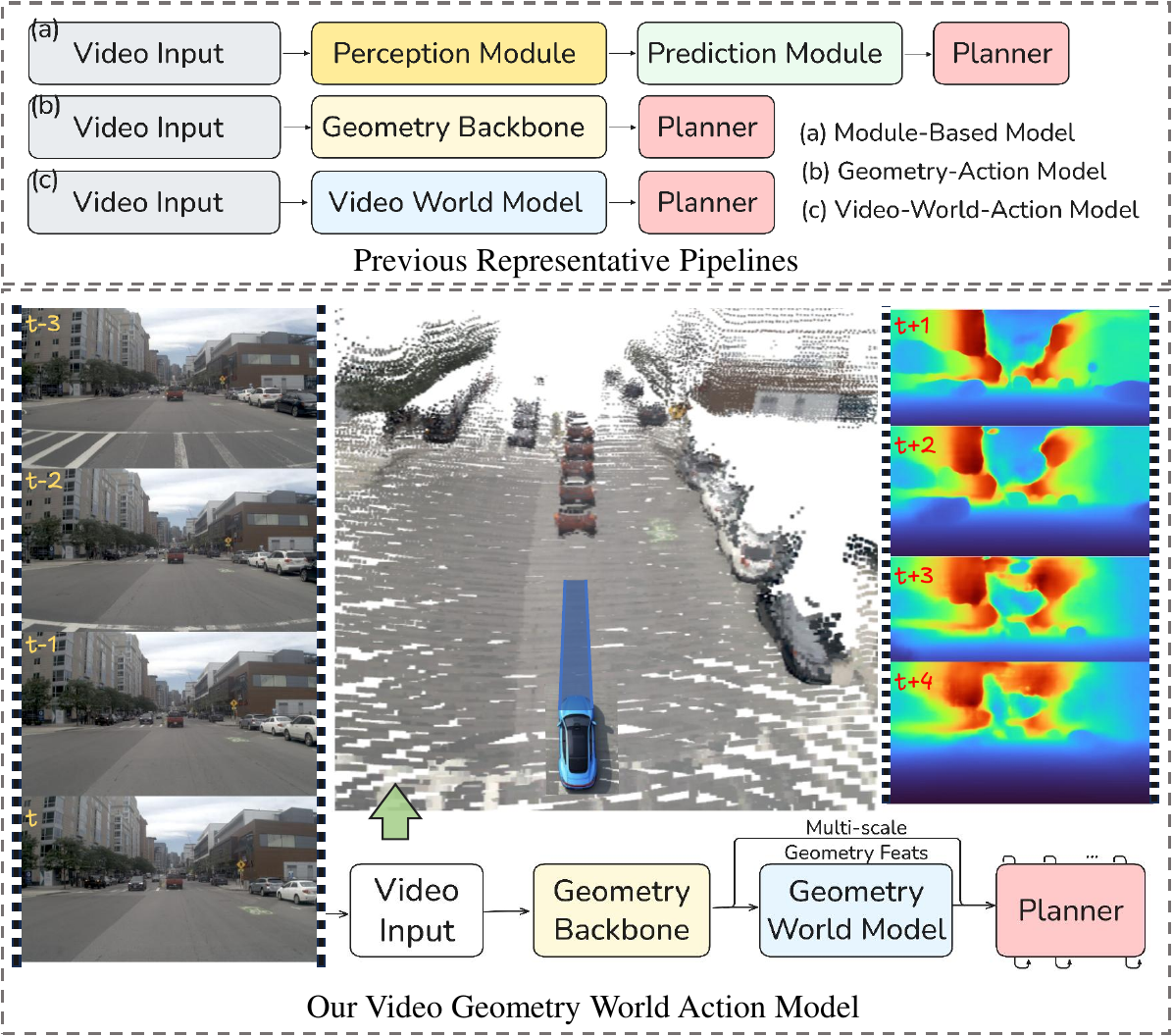}
\caption{An intuitive comparison between our video geometry world action model and previous representative pipelines. Given a consecutive video input, our \OURS provides progressively optimized trajectory planning based on the present and future geometry guidance. }
\vspace{-3mm}

\label{fig:teaser}
\end{figure*}

Autonomous driving requires safe and efficient planning in complex, open-world environments. Conventional systems decompose the driving stack into perception, prediction, planning, and control modules~\citep{vad, uniad, vadv2, stp3, paradrive, ipad} as demonstrated in Fig.~\ref{fig:teaser} (a). Although such modular pipelines improve interpretability, their hand-crafted interfaces may introduce compounding errors and limit scalability. Recent Vision/Video-Action (VA) models learn action-relevant representations directly from visual observations and scale well with advances in representative vision transformer architectures~\citep{clipvit, dinov2, dino} and large-scale training data ~\citep{openscene, waymo, nuscenes}, showing promising results in autonomous driving~\citep{drivetransformer, wisead, openread, automot, autovla, recogdrive}. Despite this progress, trajectory planning is not merely visual action prediction. It is therefore essential to ground reliable driving decisions in 3D scene geometry, including but not limited to road layout, drivable space, obstacles, dynamic agents, and their spatial relationships, which provide critical spatial cues for avoiding potential collision risks.

Pioneering work~\citep{dvgtv2} builds the planner on top of a single-layer geometry feature, introducing explicit geometric priors into policy learning as represented in Fig.~\ref{fig:teaser} (b). However, a single geometry layer may struggle to capture the diverse spatial cues required for planning: fine-grained geometry features are helpful for depicting obstacle boundaries and drivable areas, while higher-level geometry features can encode broader scene structure and agent layout. This raises the challenge of designing a geometry-oriented planner that efficiently aggregates multi-scale geometry guidance rather than relying on a single feature layer. Besides, geometry from current observations alone may be insufficient for dynamic driving scenes. Without future-aware guidance, the planner may behave conservatively under uncertainty, which can reduce driving efficiency and limit ego progress. Some early studies ~\citep{epona, drivelaw,drivevla-w0} use video-generation models as world models to provide future guidance in pixel space, as shown in Fig.~\ref{fig:teaser}(c). However, RGB representations are redundant and provide limited geometric guidance. Providing explicit guidance on how surrounding agents and ego-centric free space may evolve in future geometry space to support more effective trajectory planning remains an appealing yet challenging problem for autonomous driving.

To address these challenges, we propose \textbf{\OURS}, a unified video geometry world action model built upon StreamVGGT~\citep{streamvggt}. As shown in the bottom panel of Fig.~\ref{fig:teaser}, \OURS grounds trajectory planning in ego-aligned 3D geometry and provides planning guidance from both present observations and future scene evolution. Present geometry offers essential spatial constraints from the observed scene, while future geometry provides anticipatory priors on how surrounding agents and ego-centric free space may evolve. Specifically, \OURS first extracts ego-aligned, multi-scale geometry tokens using a streaming video geometry foundation model. A Q-Former-style geometry world model subsequently learns latent future geometry tokens conditioned on ego states, with future depth prediction providing supervision for capturing short-horizon geometric evolution. Finally, a geometry-oriented action model progressively aggregates multi-scale present geometry and latent future geometry through iterative trajectory refinement, supporting safe and efficient planning. Fig.~\ref{fig:teaser} presents an intuitive comparison between \OURS and representative planning paradigms.
Our contributions are summarized as follows:
\begin{enumerate}
    \item We formulate autonomous driving planning as a \emph{geometry world action} problem and propose \OURS, which grounds trajectory planning in explicit, ego-aligned 3D geometry and anticipates short-horizon scene evolution in the same geometric space.
    
    \item We establish present and future geometry as complementary planning guidance: multi-scale present geometry provides spatial constraints for safe motion, while latent future geometry anticipates agent and free-space evolution to support efficient driving progress.
    
    \item Extensive experiments validate the effectiveness of our \OURS, which achieves the state-of-the-art performance on NAVSIM v1 and NAVSIM v2 benchmarks.
\end{enumerate}
\section{Related Work}

Recent end-to-end autonomous driving systems can be broadly grouped into Vision-Action and Video-Action models. Vision-Action models jointly optimize perception, prediction, and planning within a differentiable framework, mapping visual observations to driving decisions. Representative methods include UniAD~\citep{uniad}, VAD~\citep{vad}, VADv2~\citep{vadv2}, SparseDrive~\citep{sparsedrive}, WoTE~\citep{wote}, DiffusionDrive~\citep{diffusiondrive}, and DriveSuprim~\citep{drivesuprim}. These methods improve planning through task-specific queries, structured scene representations, sparse or diffusion-based trajectory generation, and learned proposal evaluation. Despite their strong performance, many rely on structured supervision, such as object annotations, maps, drivable areas, and expert trajectories. Moreover, although some explicitly predict future agents or scene states, future 3D geometry is rarely treated as a primary representation for planning.

Transformer-based Video-Action models further exploit temporal context and benefit from scaling model capacity and training data~\citep{autovla, wisead, recogdrive, adathinkdrive}. Recent studies additionally incorporate world modeling to anticipate scene evolution from driving videos. Representative methods include DriveVLA-W0~\citep{drivevla-w0}, LFG~\citep{lfg}, Epona~\citep{epona}, DriveLaW~\citep{drivelaw}, and WorldDrive~\citep{worlddrive}. They improve temporal reasoning through future image generation, self-supervised video learning, diffusion-based rollouts, shared world-planning latent spaces, or trajectory-conditioned prediction. However, many of these approaches learn future guidance primarily in pixel or appearance-oriented latent spaces, with geometry serving as auxiliary supervision rather than the central planning representation.

EponaV2~\citep{eponav2} takes a step toward geometry-aware anticipation by predicting future semantic and depth maps. Nevertheless, its planning representation is primarily built upon Qwen3-VL features and lacks explicit present-scene grounding from a geometry foundation model, potentially limiting the spatial specificity of its current-scene guidance. In contrast, \OURS builds upon a video geometry foundation model and adopts ego-aligned present and future geometry as the primary planning representation, providing compact, spatially explicit, and action-relevant 3D guidance for trajectory generation.

\subsection{Geometry Foundation Models}
Recent feed-forward reconstruction methods replace per-scene optimization with direct prediction of 3D representations from sparse or multi-view images, including implicit fields, meshes, Gaussian primitives, radiance fields, and dense point maps~\cite{hong2023lrm,wang2023pflrm,li2023instant3d,wang2024crm,tochilkin2024triposr,xu2024instantmesh,xu2024grm,tang2024lgm,zhang2024gslrm,szymanowicz2024splatter,gao2024cat3d,charatan2024pixelsplat,chen2024mvsplat,chen2024lara,ye2024noposplat,smart2024splatt3r,jiang2025anysplat}. Dense coordinate regression has become particularly influential, from CroCo~\cite{weinzaepfel2022croco}, DUSt3R~\cite{dust3r}, and MASt3R~\cite{leroy2024mast3r} to extensions for dynamic scenes, sparse uncalibrated views, large view sets, and efficient inference~\cite{monst3r,zhang2025flare,tang2025mvdust3r,cabon2025must3r,yang2025fast3r,vggt,wang2026vggtomega}. However, many of these methods still process pairwise, sparse, or full-batch inputs, causing synchronization overhead or quadratic attention growth on long videos. To address this, streaming reconstruction updates geometry online as frames arrive. Existing systems use recurrent tracking, local optimization, neural implicit maps, explicit Gaussian maps, or persistent memory mechanisms for online 3D perception~\cite{teed2021droid,teed2023dpvo,zhu2022niceslam,wang2023coslam,keetha2024splatam,matsuki2024gaussiansplatting,murai2025mast3rslam,liu2025slam3r,wang2025spann3r,wang2025cut3r,wu2025point3r,chen2025long3r,pomato}. Recent transformer-based methods further adopt causal attention and cached historical states, with newer variants exploring bounded or compressed KV caches for long streams~\cite{streamvggt,lan2025stream3r,yuan2026infinitevggt,lu2026ovggt,liu2026streamcachevggt}. In autonomous driving, ~\citep{dvgt, dvgtv2} explore the efficient 4D reconstruction tailored for outdoor scenarios.

\section{Method}
\begin{figure*}[!t]
\centering
\includegraphics[width=1.0\textwidth]{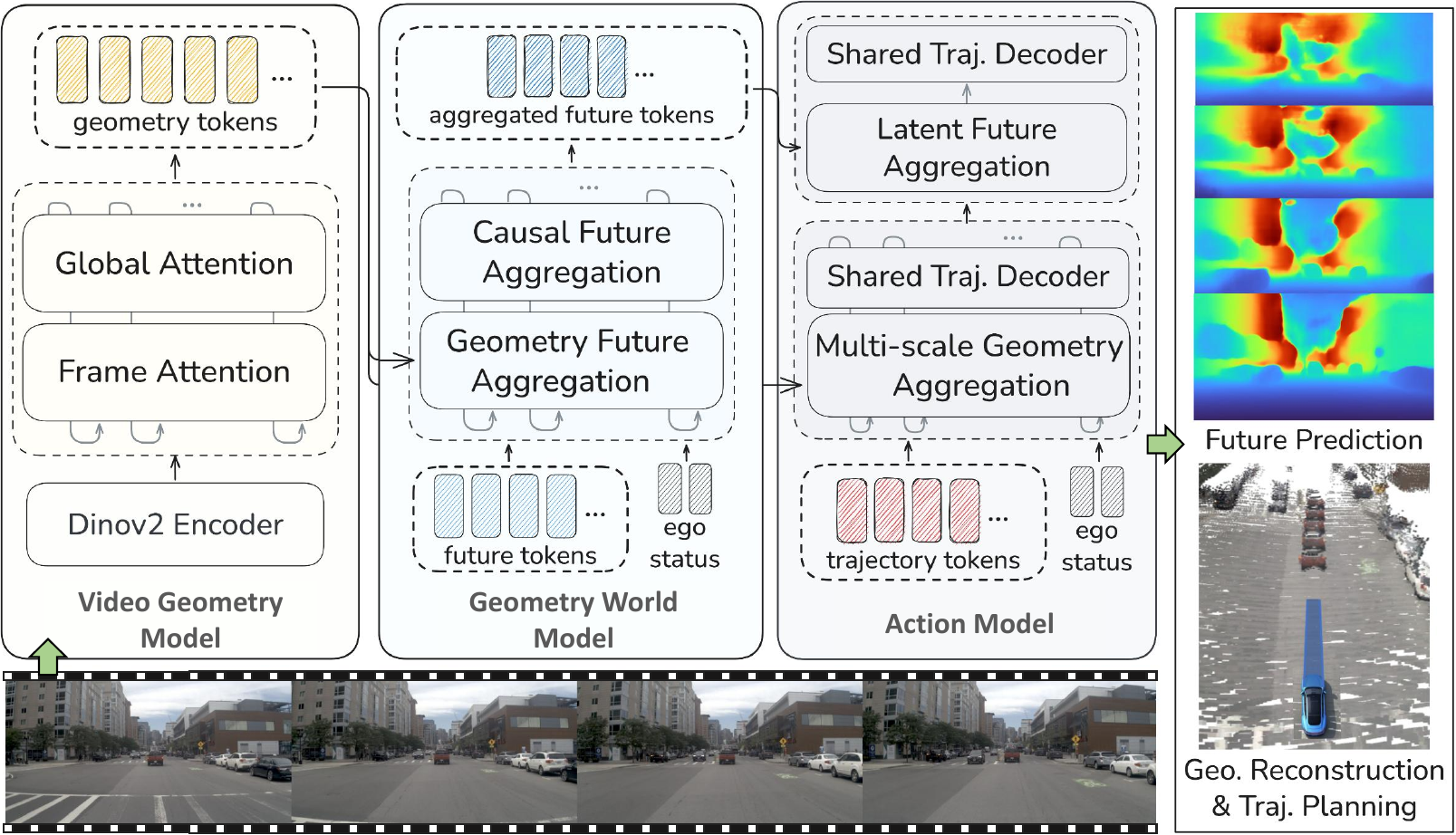}
\caption{Overview of the \OURS framework. Given an input video sequence, \OURS integrates a video geometry model, a geometry world model, and a geometry-oriented action model for 4D scene reconstruction, future depth estimation, and trajectory planning, respectively. \OURS grounds trajectory planning in ego-aligned present geometry and latent future geometry, providing both current spatial constraints and future-aware guidance. The decoders of the video geometry and geometry world models are omitted for clarity.}
\label{fig:framework}
\vspace{-3mm}
\end{figure*}

As illustrated in Fig.~\ref{fig:framework}, \OURS consists of three components: a video geometry model, a geometry world model, and a geometry-conditioned action model. Given an input video sequence, the video geometry model extracts spatio-temporal geometric representations, the geometry world model predicts short-term geometric evolution with latent future tokens, and the action model aggregates present and future geometry to generate reliable and safe trajectories. 

\subsection{Video Geometry Model}
Our video geometry model builds upon StreamVGGT~\citep{streamvggt}, a streaming video-based 4D geometry foundation model. Given a video sequence of $T$ frames, a DINOv2~\citep{dinov2} vision encoder maps each input frame $I_t \in \mathbb{R}^{3 \times H \times W}$ into a sequence of image patch tokens $F_t \in \mathbb{R}^{N \times C}$, where $N$ denotes the number of tokens and $C$ denotes the token dimension. A transformer-based decoder, composed of 24 blocks with frame-attention and global-attention modules, is then employed to extract spatio-temporally enhanced geometry tokens. From the intermediate features produced by the decoder, we select tokens $G_t^l$ from layers $\mathcal{L}=\{4,11,17,23\}$ to construct multi-scale geometry tokens $\mathcal{G}_t$:
\begin{align}
\mathcal{G}_t=\left(G_t^\ell\right)_{\ell\in\mathcal{L}}
=
\left(G_t^4, G_t^{11}, G_t^{17}, G_t^{23}\right).
\end{align}\label{eq:multiscale_geo}
These multi-scale geometry tokens are fed into DPT heads~\citep{dpt} to predict point map $P_t \in \mathbb{R}^{3 \times H \times W}$ and depth map $D_t \in \mathbb{R}^{H \times W}$. Camera parameters $g_t \in \mathbb{R}^{9}$ are estimated by introducing independent camera tokens that interact with the geometry tokens through iterative geometry decoders. In the standard setting, the predicted sequence of point maps $P_t$ is aligned to the anchor coordinate system of the first frame, while the camera pose $g_t$ represents the relative transformation from timestep $t$ to the first frame. Such a shared coordinate system enables feed-forward streaming 3D reconstruction.

For autonomous driving, grounding trajectory planning in ego-centric 3D geometry requires the geometry and trajectory representations to share a consistent coordinate system. However, StreamVGGT reconstructs scene geometry in a fixed reference frame, whereas planning trajectories are expressed in the moving ego frame, leading to increasing spatial misalignment over time. To address this issue, we express each point map in the ego-camera coordinate system of its corresponding timestep and represent camera poses as relative transformations between adjacent frames. We term this ego-aligned variant \textbf{EgoStreamVGGT}, which provides spatially consistent geometry tokens for downstream trajectory planning. The loss functions for 4D reconstruction follow the setting in StreamVGGT:
\begin{align}
L_{\mathrm{recon}} = L_\mathrm{camera} + L_\mathrm{depth} + L_\mathrm{pmap}.
\label{eq:rec_loss}
\end{align}
The details of each loss function are provided in the supplementary material.

\subsection{Geometry World Model}
The architecture of our proposed geometry world model is illustrated in the center of Fig.~\ref{fig:framework}. We maintain a set of learnable latent future tokens $Q_{\mathrm{fut}} \in \mathbb{R}^{K \times M \times C}$, where $K=4$ denotes the number of future chunks for 2 seconds, $M=64$ is the number of latent tokens per chunk. To distinguish different future horizons, we add a learnable temporal embedding to each future chunk. In addition, the ego status, including vehicle velocity, steering state, and high-level driving command, is projected by an MLP into ego embeddings $E_{\mathrm{ego}}$, which are concatenated with the geometry tokens to provide motion-state context. To extract future-aware latent geometry representations for trajectory planning, we introduce a Q-Former-style module with four geometry-guided aggregation stages, corresponding to the selected geometry layers $\mathcal{L}=\{4,11,17,23\}$. Each stage consists of a geometry-future aggregation block followed by a causal future aggregation block. For each selected layer $\ell\in\mathcal{L}$, the future tokens first cross-attend to the present geometry tokens, aggregating spatial-temporal information from the observed scene, which are then updated via causal temporal self-attention in the causal future aggregation block, where the latent future tokens of each future chunk can only attend to themselves and preceding chunks. This causal design preserves the temporal dependency of future predictions. The update process for latent future queries $Q_\mathrm{fut}$ interacting with each layer's geometry tokens can be formulated as:
\begin{align}
Q_{\mathrm{fut}}
=
\mathrm{CausalSelfAttn}
\left(
\mathrm{CrossAttn}
\left(
Q_{\mathrm{fut}}, 
\left[ G_t^\ell ; E_{\mathrm{ego}} \right]
\right)
\right),
\end{align}
where $[\cdot;\cdot]$ represents the concatenation operation.

After interacting with all multi-scale geometry tokens $\mathcal{G}_t$, the refined latent future tokens serve as compact representations of future scene evolution. We further utilize the current geometry tokens $\mathcal{G}_t$ and the latent future tokens $Q_{\mathrm{fut}}$ as conditioning variables to predict future geometry representations:
\begin{align}
    \hat{G}_{t+k}^{\ell}
    =
    \mathrm{CrossAttn}
    \left(
    G_t^\ell, Q_{\mathrm{fut}}^{k}
    \right)
    ,
    \quad \ell \in \mathcal{L}, \; k=1,\dots,K.
\end{align}
Finally, these future geometry tokens are decoded into future depth maps $\{D_{t+k}\}_{k=1}^{4}$ sharing the same DPT depth head utilized in the video geometry model and supervised with the ground truth future depth maps $\hat D$ to compute the geometry world model loss: 
\begin{equation}
L_{\text{wm}} = \sum_{i=t+1}^{t+K} \left\| {\Sigma}_i^D \odot \left( {D}_i - \hat D_i \right) \right\| + \left\| {\Sigma}_i^D \odot \left( \nabla {D}_i - \nabla \hat D_i \right) \right\| - \alpha \log {\Sigma}_i^D,
\end{equation}
where $\odot$ denotes the element-wise product, $\nabla$ indicates the gradient, and ${\Sigma}_i^D$ is the predicted confidence map. Note that the loss of future depth estimation does not contribute to the weight update of the DPT depth head.

\subsection{Geometry World Action Model}

Given multi-scale geometry tokens \(\mathcal{G}_t\) extracted by EgoStreamVGGT and latent future tokens \(Q_{\mathrm{fut}}\) predicted by the geometry world model, we initialize a set of learnable trajectory queries $Q_{\mathrm{traj}} \in \mathbb{R}^{R \times T_p \times d},$ where \(R\) is the number of trajectory proposals, \(T_p\) is the planning horizon, and \(d\) is the embedding dimension, which are set to 64, 8, and 1024, respectively.

We first refine the trajectory queries through \(L=4\) present-geometry aggregation stages, each associated with one selected geometry scale. At each stage \(\ell\), the trajectory queries cross-attend to \(\mathcal{G}_t^\ell\) and the ego status embeddings through a transformer block. A shared MLP then decodes the updated queries into \(R\) trajectory proposals. Each proposal contains \(T_p\) future waypoints parameterized as \((x, y, \theta)\), where the heading angle \(\theta\) is constrained to \([-\pi,\pi]\) using a \(\tanh\) activation. After aggregating the present geometry, we perform an additional refinement stage using the latent future geometry tokens \(Q_{\mathrm{fut}}\). Specifically, the trajectory queries attend to \(Q_{\mathrm{fut}}\) through another transformer block, allowing the proposals to incorporate anticipatory guidance about short-horizon scene evolution.

The complete refinement process produces a sequence of trajectory predictions
\(\{P^{(j)}\}_{j=1}^{N_{\mathrm{ref}}}\), where
\(N_{\mathrm{ref}}=5\), comprising four present-geometry aggregation stages and one future-geometry refinement stage. We supervise each stage using a minimum-over-proposals objective. Let
\(P_r^{(j)}\in\mathbb{R}^{T_p\times3}\) denote the \(r\)-th proposal at stage \(j\), and let
\(\hat{P}_{\mathrm{gt}}\in\mathbb{R}^{T_p\times3}\) denote the ground-truth trajectory. The stage-wise trajectory loss is
\begin{align}
L_{\mathrm{traj}}^{(j)}
=
\min_{r\in\{1,\ldots,R\}}
\left\|P_r^{(j)}-\hat{P}_{\mathrm{gt}}\right\|_1 .
\end{align}
The overall trajectory loss is then computed as the weighted sum across all stages:
\begin{align}
L_{\mathrm{traj}}
=
\sum_{j=1}^{N_{\mathrm{ref}}} \lambda_j L_{\mathrm{traj}}^{(j)},    
\end{align}
where \(\lambda_j\) exponentially down-weights earlier refinement stages.

Finally, we attach a proposal-scoring head to the final-stage trajectory features. For each proposal, we pool its features along the temporal dimension and apply an MLP to predict a score \(S_r\). Following~\citep{ipad,navsim}, the target score is defined as:
\begin{align}\label{eq:pdms}
S_{\mathrm{gt}}
=
\mathrm{NC} \times \mathrm{DAC} \times \frac{5\,\mathrm{EP} + 5\,\mathrm{TTC} + 2\,\mathrm{Comf}}{12},
\end{align}

where NC, DAC, EP, TTC, and Comf denote no at-fault collision, drivable-area compliance, ego progress, time-to-collision, and comfort, respectively. All metrics are obtained using the NAVSIM simulator~\citep{navsim}. The binary cross-entropy loss $L_\mathrm{score}$ is applied to optimize the scoring head. The trajectory loss is jointly optimized with the 4D reconstruction loss $L_\mathrm{recon}$ and geometry world model loss $L_\mathrm{wm}$. 
The complete training objective for jointly optimizing the video geometry model, geometry world model, and action model is:
\begin{align}
L
=
L_{\mathrm{traj}}
+L_{\mathrm{score}}
+L_{\mathrm{recon}}
+L_{\mathrm{wm}}.
\end{align}
The auxiliary decoders for 4D reconstruction and future-depth prediction are not required during trajectory-planning inference.

\section{Experiments}

\subsection{Datasets and Evaluation Metrics}

\textbf{Datasets.} Our model is trained and evaluated on a large-scale mixture of driving data. For geometry-related tasks, we utilize real-world datasets, namely OpenScene~\citep{openscene} and nuScenes~\citep{nuscenes}, alongside synthetic data from ParallelDomain~\citep{paralleldomain} and RealDriveSim~\citep{realdrivesim}. Trajectory planning is trained and evaluated on the NAVSIM~\citep{navsim} dataset, which is a subset of representative scenarios from OpenScene. 

\textbf{Evaluation Metrics.} We adopt standard metrics to assess both planning and geometry performance. For closed-loop planning, we benchmark our model on NAVSIM v1 and v2. NAVSIM v1 simulates a 4-second non-reactive environment at 10 Hz, scoring agents via the Predictive Driver Model Score (PDMS), which aggregates fundamental safety, comfort, and progress metrics, as defined in Eq.~\ref{eq:pdms}. 
NAVSIM v2 enhances simulation realism with reactive traffic and introduces the Extended PDMS (EPDMS), which incorporates supplementary criteria such as traffic rule compliance. We also report video depth evaluation and camera pose estimation to demonstrate the geometry performance of our modified EgoStreamVGGT following the evaluation protocol of StreamVGGT. Camera pose evaluation metrics are provided in the supplementary material.

\subsection{Training Details}

The training pipeline for \OURS proceeds in three distinct stages:

\textbf{Stage 1:} We first pretrain EgoStreamVGGT on a mixture of synthetic and real-world datasets, including OpenScene~\citep{openscene}, nuScenes~\citep{nuscenes}, ParallelDomain~\citep{paralleldomain}, and RealDriveSim~\citep{realdrivesim}. The model is initialized from a pretrained StreamVGGT checkpoint and optimized for 23K steps. The sampling ratio among the four datasets is set to \(10{:}10{:}1{:}1\), respectively.

\textbf{Stage 2:} Starting from the Stage 1 checkpoint, we train two branches in parallel. First, the geometry world model learns latent future geometry tokens under future-depth supervision while retaining the 4D reconstruction objectives. Given four consecutive input frames, it predicts the depth maps of four future frames spanning the subsequent two seconds. This branch is trained on OpenScene for 47K steps. Meanwhile, we train the trajectory planner to aggregate multi-scale present geometry tokens. Using the standard \textit{navtrain} split of NAVSIM~\citep{navsim}, the planner takes four consecutive frames as input and predicts eight future waypoints over a four-second horizon. This branch is trained for 32K steps, resulting in an intermediate model termed \textbf{GeoAD}. GeoAD grounds planning in present-scene geometry without incorporating future geometry guidance. 

\textbf{Stage 3:} We construct the complete \OURS model by integrating the GeoAD planner with the geometry world model obtained in Stage 2. To preserve the initial planning behavior of GeoAD, we zero-initialize the output projection of the future-geometry aggregation block. Consequently, this block initially produces zero residual corrections and gradually learns to refine the trajectory proposals using future geometry guidance. The complete model is trained on the NAVSIM \textit{navtrain} split for an additional 64K steps.

Across all stages, we use the AdamW optimizer with a global batch size of 64 distributed across 32 NVIDIA H20 GPUs. The learning rate is set to \(1\times10^{-4}\) in Stages 1 and 2 and \(1\times10^{-5}\) in Stage 3, following a cosine learning-rate schedule.

\subsection{Experimental Results}
\begin{table*}[t]
\centering
\label{tab:navsim_closedloop}
\renewcommand{\arraystretch}{1.1}
\setlength{\tabcolsep}{4pt}
\resizebox{0.95\textwidth}{!}{
\begin{tabular}{l|c|c|cccccc}
\toprule

\textbf{Method} 
& \textbf{Input} 
& \textbf{Aux. Sup.} 
& \textbf{NC$\uparrow$} 
& \textbf{DAC$\uparrow$} 
& \textbf{TTC$\uparrow$} 
& \textbf{Comf.$\uparrow$} 
& \textbf{EP$\uparrow$} 
& \textbf{PDMS$\uparrow$} \\

\midrule

\color{gray!100} PARA-Drive~\citep{paradrive} & \color{gray!100} C & \color{gray!100} Map \& Mot. \& Occ & \color{gray!100} 97.9 & \color{gray!100} 92.4 & \color{gray!100} 93.0 & \color{gray!100} 99.8 & \color{gray!100} 79.3 & \color{gray!100} 84.0 \\

\color{gray!100} VADv2~\citep{vadv2} & \color{gray!100} C & \color{gray!100} Map \& Mot. \& Traffic & \color{gray!100} 97.2 & \color{gray!100} 89.1 & \color{gray!100} 91.6 & \color{gray!100} 100 & \color{gray!100} 76.0 & \color{gray!100} 80.9 \\

\color{gray!100} UniAD~\citep{uniad} & \color{gray!100} C & \color{gray!100} Map \& Box \& Mot. \& Occ & \color{gray!100} 97.8 & \color{gray!100} 91.9 & \color{gray!100} 92.9 & \color{gray!100} 100 & \color{gray!100} 78.8 & \color{gray!100} 83.4 \\

\color{gray!100} iPad~\citep{ipad} & \color{gray!100} C & \color{gray!100} Map \& Box & \color{gray!100} 98.6 & \color{gray!100} 98.3 & \color{gray!100} 94.9 & \color{gray!100} 100 & \color{gray!100} 88.0 & \color{gray!100} 91.7 \\

\color{gray!100} Transfuser~\citep{transfuser} & \color{gray!100} C \& L & \color{gray!100} Map \& Box & \color{gray!100} 97.7 & \color{gray!100} 92.8 & \color{gray!100} 92.8 & \color{gray!100} 100 & \color{gray!100} 79.2 & \color{gray!100} 84.0 \\

\color{gray!100} GoalFlow~\citep{goalflow} & \color{gray!100} C \& L & \color{gray!100} Map \& Box & \color{gray!100} 98.3 & \color{gray!100} 93.8 & \color{gray!100} 94.3 & \color{gray!100} 100 & \color{gray!100} 79.8 & \color{gray!100} 85.7 \\

\color{gray!100} DiffusionDrive~\citep{diffusiondrive} & \color{gray!100} C \& L & \color{gray!100} Map \& Box & \color{gray!100} 98.2 & \color{gray!100} 96.2 & \color{gray!100} 94.7 & \color{gray!100} 100 & \color{gray!100} 82.2 & \color{gray!100} 88.1 \\

\color{gray!100} WoTE~\citep{wote} & \color{gray!100} C \& L & \color{gray!100} Map \& Box & \color{gray!100} 98.5 & \color{gray!100} 96.8 & \color{gray!100} 94.9 & \color{gray!100} 99.9 & \color{gray!100} 81.9 & \color{gray!100} 88.3 \\

\color{gray!100} DriveSuprim~\citep{drivesuprim} & \color{gray!100} C \& L & \color{gray!100} Map \& Box & \color{gray!100} 97.8 & \color{gray!100} 97.3 & \color{gray!100} 93.6 & \color{gray!100} 100 & \color{gray!100} 86.7 & \color{gray!100} 89.9 \\
\midrule

Epona~\citep{epona} & C &  Future States & 97.9 & 95.1 & 93.8 & 99.9 & 80.4 & 86.2 \\
DriveLaW~\citep{drivelaw} & C & Future States & 99.0 & 97.1 & \textbf{96.7} & \textbf{100} & 81.3 & 89.1 \\
WorldDrive~\citep{worlddrive} & C & Future States & 98.4 & 96.8 & 95.2 & \textbf{100} & 83.3 & 89.0 \\
DriveVLA-W0~\citep{drivevla-w0} & C & Future States& 98.7 & \textbf{99.1} & 95.3 & 99.3 & 83.3 & 90.2 \\
EponaV2~\citep{eponav2} & C & Future States & 98.6 & 97.9 & 95.7 & \textbf{100} & 84.8 & 90.4 \\ \midrule

LFG~\citep{lfg} & C & Dense Geometry & 98.2 & 93.7 & 94.4 & \textbf{100} & 79.1 & 85.2 \\
DVGT-2~\citep{dvgtv2} 
& C 
& Dense Geometry 
& 98.7 & 97.9 & 95.8 & \textbf{100} & 84.3 & 90.3 \\
\midrule
\rowcolor{gray!15}\OURS(OURS) & C & Dense \& Future Geo. & \textbf{99.0} & 97.8  & 95.8 & 99.9 & \textbf{85.9} & \textbf{91.0} \\

\bottomrule
\end{tabular}
}
\vspace{1mm}
\caption{
Closed-loop planning results on NAVSIM v1 navtest split. C and L are short for camera and lidar. Aux. Sup. is short for auxiliary supervision. Our GeoWorldAD achieves the state-of-the-art PDMS metric among competing world-model based and geometry-based methods. 
}\label{tab:nav_v1}
\vspace{-2mm}
\end{table*}

\begin{table}[t]
\centering
\label{tab:navsim_v2_navtest}
\resizebox{0.95\textwidth}{!}{
\begin{tabular}{l|ccccccccc|c}
\toprule
\textbf{Method} & \textbf{NC} $\uparrow$ & \textbf{DAC} $\uparrow$ & \textbf{DDC} $\uparrow$ & \textbf{TL} $\uparrow$ & \textbf{EP} $\uparrow$ & \textbf{TTC} $\uparrow$ & \textbf{LK} $\uparrow$ & \textbf{HC} $\uparrow$ & \textbf{EC} $\uparrow$ & \textbf{EPDMS} $\uparrow$ \\
\midrule

\color{gray!100} Transfuser~\cite{transfuser} & \color{gray!100} 96.9 & \color{gray!100} 89.9 & \color{gray!100} 97.8 & \color{gray!100} 99.7 & \color{gray!100} 87.1 & \color{gray!100} 95.4 & \color{gray!100} 92.7 & \color{gray!100} 98.3 & \color{gray!100} 87.2 & \color{gray!100} 76.7 \\

\color{gray!100} DriveSuprim~\cite{drivesuprim} & \color{gray!100} 97.5 & \color{gray!100} 96.5 & \color{gray!100} 99.4 & \color{gray!100} 99.6 & \color{gray!100} 88.4 & \color{gray!100} 96.6 & \color{gray!100} 95.5 & \color{gray!100} 98.3 & \color{gray!100} 77.0 & \color{gray!100} 83.1 \\

\color{gray!100} DiffusionDrive~\cite{diffusiondrive} & \color{gray!100} 98.2 & \color{gray!100} 95.9 & \color{gray!100} 99.4 & \color{gray!100} 99.8 & \color{gray!100} 87.5 & \color{gray!100} 97.3 & \color{gray!100} 96.8 & \color{gray!100} 98.3 & \color{gray!100} 87.7 & \color{gray!100} 84.5 \\
\midrule
DriveVLA-W0~\citep{drivevla-w0}
                     & 98.5 & \textbf{99.1} & 98.0 & 99.7 & 86.4 & 98.1 & 93.2 & 97.9 & 58.9 & 86.1 \\
DVGT-2~\citep{dvgtv2}
                     & 98.7 & 97.9 & \textbf{99.7} & \textbf{99.9} & 87.9 & 98.0 & \textbf{98.2} & \textbf{98.2} & 77.0 & 89.6 \\
EponaV2~\citep{eponav2}
                     & 98.5 & 97.4 & 99.5 & \textbf{99.9} & 87.9 & 98.1 & 97.7 & 98.2 & 77.4 & 88.9 \\
\midrule
\rowcolor{gray!15}\OURS(OURS)    
                     & \textbf{99.0} & 97.8 & 99.6 & 99.7 & \textbf{89.1} & \textbf{98.6} & 97.6 
                     & 98.0 & \textbf{82.2} & \textbf{90.4}
\\
\bottomrule
\end{tabular}
}
\vspace{2mm}
\caption{Closed-loop planning results on NAVSIM v2 navtest split. Our \OURS achieves the best EPDMS driving score among all the competing methods.}\label{tab:nav_v2}
\vspace{-6mm}
\end{table}

\subsubsection{Closed-Loop Evaluation on NAVSIM Benchmarks}

We report closed-loop planning results on NAVSIM v1 and v2 in Tab.~\ref{tab:nav_v1} and~\ref{tab:nav_v2}, respectively. The compared methods include perception-based pipelines with structured supervision, world-model-based planners using future-state prediction, and geometry-oriented planners using dense present-scene geometry. \OURS achieves the best performance among perception-free methods on NAVSIM v1 and the highest overall EPDMS on NAVSIM v2.

On NAVSIM v1, \OURS achieves a PDMS of 91.0, outperforming the strongest dense-geometry baseline, DVGT-2 (90.3), and the strongest future-state-based method, EponaV2 (90.4). Compared with DVGT-2, \OURS improves EP from 84.3 to 85.9 while increasing NC from 98.7 to 99.0 and maintaining the same TTC of 95.8, suggesting that future geometry guidance improves driving progress without compromising safety. Compared with EponaV2, which introduces future semantic and depth prediction as auxiliary supervision, \OURS improves PDMS by 0.6 points and EP by 1.1 points, while also achieving higher NC and TTC. This comparison highlights the benefit of grounding trajectory planning in explicit 3D geometry that captures both the observed scene and its anticipated future evolution.

On NAVSIM v2, \OURS achieves the highest EPDMS of 90.4, exceeding DVGT-2 and EponaV2 by 0.8 and 1.5 points, respectively. It also improves the key safety and efficiency metrics over both methods, achieving 99.0 NC, 98.6 TTC, and 89.1 EP. Overall, these results demonstrate the complementary roles of dense present geometry and latent future geometry: the former provides explicit spatial constraints, while the latter anticipates scene evolution, enabling \OURS to better balance driving safety and progress.

\begin{table*}[t]
\centering
\renewcommand{\arraystretch}{1.1}
\setlength{\tabcolsep}{4pt}
\resizebox{0.75\textwidth}{!}{
\begin{tabular}{l|cccc|cccc}
\toprule
\multirow{2}{*}{\textbf{Method}} & \multicolumn{4}{c|}{\textbf{v1}} & \multicolumn{4}{c}{\textbf{v2}} \\
& \textbf{NC$\uparrow$} 
& \textbf{TTC$\uparrow$} 
& \textbf{EP$\uparrow$} 
& \textbf{PDMS$\uparrow$}
& \textbf{NC$\uparrow$} 
& \textbf{TTC$\uparrow$} 
& \textbf{EP$\uparrow$} 
& \textbf{EPDMS$\uparrow$} \\
\midrule

GeoAD & 98.9 & 95.7 & 82.6 & 89.3 & 98.9 & 98.3 & 86.3 & 87.6 \\

\OURS & \textbf{99.0} & \textbf{95.8} & \textbf{85.9} & \textbf{91.0} & \textbf{99.0} & \textbf{98.6} & \textbf{89.1} & \textbf{90.4} \\

\bottomrule
\end{tabular}
}
\caption{
Ablation study on latent future geometry tokens. Incorporating future geometry anticipation improves ego progress and slightly enhances safety performance.
}\label{tab:wm_ablation}
\vspace{-3mm}
\end{table*}
\begin{table*}[t]
\centering

\renewcommand{\arraystretch}{1.1}
\setlength{\tabcolsep}{4pt}
\begin{tabular}{l|c|cccccc}
\toprule

\textbf{Pretrained Model} 
& \textbf{Aux. Sup.} 
& \textbf{NC$\uparrow$} 
& \textbf{DAC$\uparrow$} 
& \textbf{TTC$\uparrow$} 
& \textbf{Comf.$\uparrow$} 
& \textbf{EP$\uparrow$} 
& \textbf{PDMS$\uparrow$} \\

\midrule

Scratch & - & 98.1 & 94.6 & 93.9 & 99.1 & 76.0 & 84.2 \\
StreamVGGT & 4D Recon. & 97.9 & 93.4 & 92.8 & 99.8 & 80.2 &
84.8 \\
EgoStreamVGGT & - & 98.4 & 95.1 & 95.0 & 99.9 & 81.7 &
87.3 \\
EgoStreamVGGT & 4D Recon. & \textbf{98.9} & \textbf{97.2}  & \textbf{95.7} & \textbf{99.9} & \textbf{82.6} & \textbf{89.3} \\

\bottomrule
\end{tabular}
\caption{
Ablation study on the effectiveness of geometry representation on the NAVSIM v1 navtest split. Ego-aligned representation and joint geometric supervision introduce consistent improvement.
}\label{tab:geo_rep}
\vspace{-2mm}
\end{table*}

\subsubsection{Ablation Study on the Geometry World Model}


We evaluate the effectiveness of the geometry world model in Tab.~\ref{tab:wm_ablation} by comparing GeoAD with the complete GeoWorldAD. GeoAD performs planning solely with multi-scale present geometry, while keeping the remaining components unchanged. It already achieves an NC score of 98.9 on both NAVSIM v1 and v2, indicating that explicit present geometry provides a strong spatial foundation for safety-critical planning. Incorporating latent future geometry tokens consistently improves all reported metrics. On NAVSIM v1, GeoWorldAD improves NC from 98.9 to 99.0, TTC from 95.7 to 95.8, EP from 82.6 to 85.9, and PDMS from 89.3 to 91.0. On NAVSIM v2, it improves NC from 98.9 to 99.0, TTC from 98.3 to 98.6, EP from 86.3 to 89.1, and EPDMS from 87.6 to 90.4. Notably, the largest gains are observed in ego progress, with improvements of 3.3 and 2.8 points on v1 and v2, respectively, while the safety-related metrics are maintained or slightly improved.

These results highlight the central advantage of GeoWorldAD over present-geometry-only planning. Present geometry provides explicit spatial constraints from the observed scene, whereas future geometry anticipates how surrounding agents and ego-centric free space may evolve. Their combination enables more progressive planning under uncertainty without sacrificing safety, leading to a better balance between collision avoidance and driving efficiency.

\subsubsection{Ablation Study on Geometry Representation and Supervision}

We investigate the effects of geometry representation and joint 4D reconstruction supervision in Tab.~\ref{tab:geo_rep}. To isolate these factors, all variants perform planning using only multi-scale present geometry tokens, without latent future geometry tokens. Training the planner from scratch establishes a baseline PDMS of 84.2. Introducing vanilla StreamVGGT with 4D reconstruction supervision improves EP from 76.0 to 80.2, but yields only a marginal PDMS gain of 0.6 and decreases NC, DAC, and TTC. This mixed result suggests that geometry pretraining alone does not necessarily provide effective planning guidance when its fixed-frame representation is misaligned with the ego-centric coordinate system used for trajectory prediction.

EgoStreamVGGT addresses this discrepancy by representing point maps in their ego-camera frames and camera poses as relative transformations between adjacent frames. Even without auxiliary 4D reconstruction supervision, EgoStreamVGGT improves PDMS from 84.8 to 87.3 over vanilla StreamVGGT, together with consistent gains in NC, DAC, TTC, and EP. This result validates the importance of grounding trajectory planning in coordinate-consistent, ego-centric 3D geometry.

Jointly optimizing the 4D reconstruction objective further improves EgoStreamVGGT across all planning metrics. These gains indicate that additional geometric supervision helps preserve spatially informative representations during planner training, further enhancing driving safety.

Overall, the results show that effective geometry-oriented planning depends on both an ego-aligned representation and joint geometric supervision, rather than geometry pretraining alone. This configuration forms our present-geometry planner, GeoAD, which is further extended with latent future geometry tokens in GeoWorldAD. The corresponding video depth evaluation is reported in Tab.~\ref{tab:depth_eval_main}. Detailed experiment analysis on 4D reconstruction can be found in the supplementary materials.

\begin{table*}[!t]
\centering
\renewcommand{\arraystretch}{1.1}
\setlength{\tabcolsep}{4pt}
\resizebox{0.95\textwidth}{!}{
\begin{tabular}{l l|cc|cc|cc}
\toprule
& & \multicolumn{2}{c|}{\textbf{OpenScene}} 
& \multicolumn{2}{c|}{\textbf{nuScenes}} 
& \multicolumn{2}{c}{\textbf{KITTI}} \\

\textbf{Method} & \textbf{Type} 
& \textbf{Abs Rel$\downarrow$} & $\boldsymbol{\delta < 1.25 \uparrow}$ 
& \textbf{Abs Rel$\downarrow$} & $\boldsymbol{\delta < 1.25 \uparrow}$ 
& \textbf{Abs Rel$\downarrow$} & $\boldsymbol{\delta < 1.25 \uparrow}$ \\
\midrule

StreamVGGT & Streaming & 0.236 & 65.6 & 0.265 & 58.2 & 0.173 & 72.2 
 \\
EgoStreamVGGT & Streaming & \textbf{0.141} & \textbf{86.5} & \textbf{0.117} & \textbf{88.5} & \textbf{0.077} & \textbf{95.5} \\
\bottomrule
\end{tabular}
}
\caption{Video depth estimation. Our finetuned EgoStreamVGGT achieves consistent improvements across OpenScene, nuScenes, and KITTI datasets.}\label{tab:depth_eval_main}
\vspace{-3mm}
\end{table*}
\section{Limitations}
Although EgoStreamVGGT supports continuous video streams for 4D reconstruction, the current trajectory planner operates on fixed-length clips. Integrating KV caching to enable efficient streaming inference for trajectory planning remains a promising direction for future work.
\section{Conclusion}

We propose \OURS, a unified video geometry world action model that grounds autonomous driving planning in ego-aligned 3D geometry. \OURS combines multi-scale present geometry with latent future geometry modeling, providing complementary spatial constraints and anticipatory guidance for trajectory planning. Its geometry-oriented action model progressively aggregates these geometric cues through iterative trajectory refinement, improving driving progress while preserving safety. Extensive experiments on NAVSIM v1 and v2 demonstrate state-of-the-art closed-loop performance and validate the importance of coordinate-consistent geometry grounding and future geometry anticipation for safe and efficient planning.




\bibliography{example}  

\newpage
\appendix
\onecolumn

\section{Appendix for GeoWorldAD}
\subsection{Video Geometry Reconstruction Losses}

Following StreamVGGT, EgoStreamVGGT is trained with three reconstruction objectives: a camera loss \(L_{\mathrm{camera}}\), a depth loss \(L_{\mathrm{depth}}\), and a point-map loss \(L_{\mathrm{pmap}}\). These objectives correspond to Eq.~\ref{eq:rec_loss} in the main paper:
\begin{align}
L_{\mathrm{recon}}
=
L_{\mathrm{camera}}
+
L_{\mathrm{depth}}
+
L_{\mathrm{pmap}}.
\end{align}

Given an input video sequence of \(T\) frames, EgoStreamVGGT predicts the camera parameters \(g_t\), depth map \(D_t\in\mathbb{R}^{H\times W}\), point map \(P_t\in\mathbb{R}^{3\times H\times W}\), and the corresponding confidence maps for each frame \(t\). Their ground-truth values are denoted by \(\hat{g}_t\), \(\hat{D}_t\), and \(\hat{P}_t\), respectively.

Unlike the anchor-frame representation adopted by StreamVGGT, EgoStreamVGGT expresses each point map \(P_t\) in the ego-camera coordinate system of its corresponding timestep. The camera parameters \(g_t\) encode the relative camera transformation between adjacent frames. This formulation retains temporal ego-motion information while keeping the reconstructed geometry aligned with the ego-centric coordinate system used for trajectory planning.

\paragraph{Camera loss.}
The camera loss supervises the predicted camera parameters using their ground-truth values. Following StreamVGGT, we adopt the Huber loss for robustness to noisy camera annotations and outliers:
\begin{align}
L_{\mathrm{camera}}
=
\sum_{t=1}^{T}
\rho_{\epsilon}\left(g_t-\hat{g}_t\right),
\end{align}
where \(\rho_{\epsilon}(\cdot)\) denotes the element-wise Huber loss. The camera parameter \(g_t\in\mathbb{R}^{9}\) contains the relative translation and rotation between adjacent frames, as well as the field-of-view parameters. This objective provides explicit supervision for relative ego-motion estimation, enabling the model to relate consecutive ego-centric geometry representations over time.

\paragraph{Depth loss.} The depth loss supervises the predicted depth map $D_t$ using the ground-truth depth $\hat{D}_t$. Since depth quality varies across pixels due to occlusion, motion, and sparse or noisy supervision, the loss is weighted by the predicted depth confidence map $\Sigma_t^D$. In addition to the depth-value error, we follow StreamVGGT and include a gradient matching term to encourage local geometric consistency and sharper depth discontinuities:

\begin{align}
L_{\mathrm{depth}} = \sum_{t=1}^{T} \left(\left| \Sigma_t^D \odot (D_t - \hat{D}_t) \right| + \left| \Sigma_t^D \odot (\nabla D_t - \nabla \hat{D}_t) \right| - \alpha \log \Sigma_t^D \right),
\end{align}

where $\odot$ denotes element-wise multiplication, $\nabla$ denotes the spatial gradient operator, and $\alpha$ controls the confidence regularization. The confidence term allows the model to down-weight uncertain regions, while the negative logarithmic regularizer prevents the confidence from collapsing to trivial low values. This loss encourages accurate dense depth prediction while preserving object boundaries and road-surface geometry.

\paragraph{Point-map loss.}
The point-map loss supervises the predicted 3D point map $P_t$ with the ground-truth point map $\hat{P}_t$. It uses the predicted point-map confidence $\Sigma_t^P$ and includes both point-coordinate and gradient-consistency terms:

\begin{align}
L_{\mathrm{pmap}} = \sum_{t=1}^{T} \left(\left| \Sigma_t^P \odot (P_t - \hat{P}_t) \right| + \left| \Sigma_t^P \odot (\nabla P_t - \nabla \hat{P}_t) \right| - \alpha \log \Sigma_t^P \right).
\end{align}

Unlike the vanilla implementation, we modify the coordinate system from the shared reference frame to the corresponding ego-camera coordinate system in EgoStreamVGGT.

\subsection{Collision-Related Metrics}

Among the closed-loop planning metrics, no-at-fault collision (NC) and time-to-collision (TTC) are most directly related to driving safety. NC measures whether the ego vehicle completes the simulated scenario without being responsible for a collision. It therefore reflects the planner's ability to respect the spatial occupancy of surrounding vehicles, pedestrians, and static obstacles under closed-loop execution. A high NC score indicates that the generated trajectory remains collision-free even after the simulator rolls out the consequences of the ego action, making it a fundamental constraint for evaluating autonomous driving safety.

TTC further complements NC by measuring the temporal margin before a potential collision. While NC captures whether a collision eventually occurs, TTC evaluates whether the ego vehicle maintains sufficient reaction time with respect to nearby dynamic agents. A low TTC usually indicates that the ego vehicle is too close to another road user, given their relative velocity, even if an actual collision has not yet happened within the simulation horizon. Therefore, TTC serves as an early-warning metric for near-collision and high-risk behaviors.

These two metrics are particularly important for GeoWorldAD because our method explicitly introduces present and future geometric priors into trajectory planning. Present geometry helps the planner understand the current drivable space and obstacle layout, while future geometry provides anticipatory cues about how the scene may evolve. By jointly optimizing with these geometric constraints, the planner can reduce both immediate collision risks, reflected by NC, and latent near-collision risks, reflected by TTC. Consequently, consistent gains in NC and TTC demonstrate that GeoWorldAD improves not only trajectory accuracy or progress, but also the safety-critical behavior required for reliable autonomous driving.

\begin{figure*}[!t]
\centering
\includegraphics[width=0.95\textwidth]{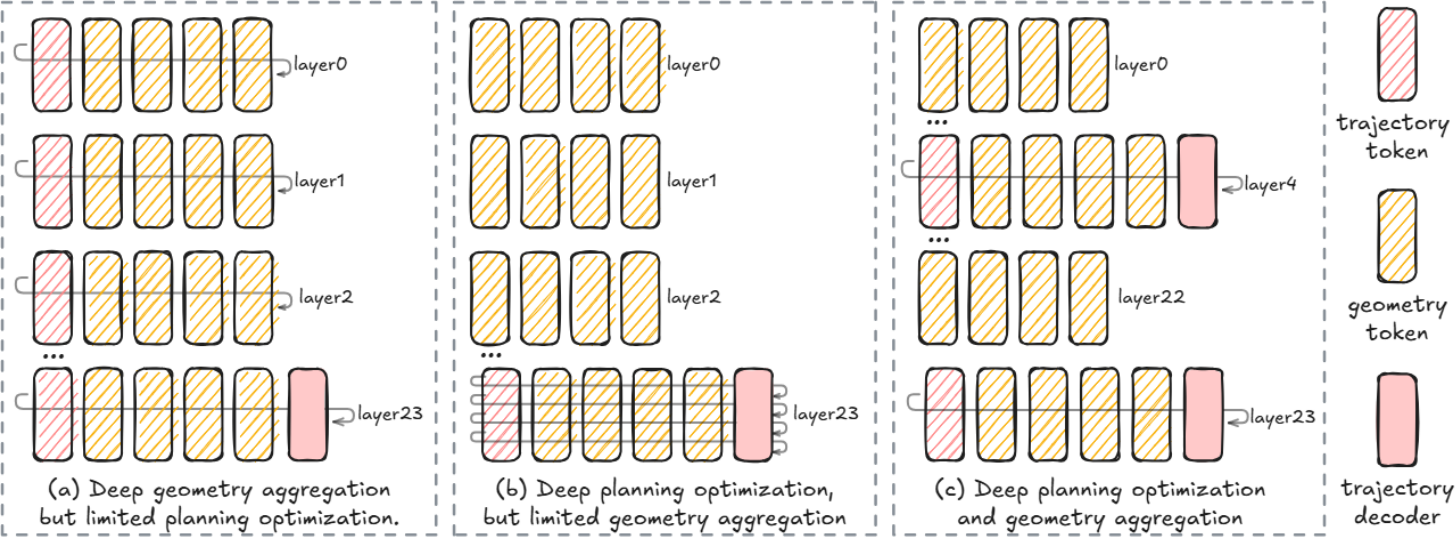}
\caption{Comparison of three different geometry aggregation strategies for trajectory planning.}
\label{fig:planning_geo}
\end{figure*}

\subsection{Additional Experiments}
\subsubsection{Exploration on Geometry Aggregation}

As shown in Fig. \ref{fig:planning_geo}, we compare different strategies for aggregating geometry priors into the trajectory planner. The first strategy, illustrated on the left of Fig. \ref{fig:planning_geo} (a), directly lets the trajectory tokens interact with the geometry tokens from all 24 layers, followed by a single trajectory decoder for supervision. This design is similar to the strategy used in ~\citep{dvgtv2}. As shown in the first row of Tab. \ref{tab:geo_plan_interaction}, although the trajectory tokens can access rich multi-layer geometry features, the planner still obtains sub-optimal performance. We attribute this to the limited optimization depth for trajectory planning: a single interaction stage is insufficient for efficiently absorbing both low-level spatial details and high-level semantic geometry cues.

To encourage more progressive interaction, we further evaluate the middle design in Fig. \ref{fig:planning_geo} (b), where the trajectory tokens are placed on top of the final-layer geometry tokens and are optimized iteratively through a shared trajectory planning decoder. This design allows the trajectory representation to be gradually refined. As shown in the second row of Tab. \ref{tab:geo_plan_interaction}, the ego-progress metric improves from 81.5 to 82.9, indicating that iterative refinement helps generate more progressive trajectories. However, because the planner only accesses the final geometry layer, the improvements in collision-related metrics remain limited. This suggests that high-level geometry alone cannot provide all the fine-grained spatial constraints required for safe planning.

Our final design, adopted in GeoWorldAD as shown in the right panel of Fig. \ref{fig:planning_geo} (c), combines the advantages of both strategies. We introduce multi-scale geometry tokens from the DPT head, which contain spatial information from shallow to deep layers, and perform iterative trajectory optimization after each geometry aggregation step. In this way, the trajectory tokens can progressively absorb geometry guidance from different representation levels. Because supervision is applied at each refinement stage, the planner learns to efficiently extract safety-critical spatial cues and convert them into collision-aware trajectory updates. As shown in the last row of Tab. \ref{tab:geo_plan_interaction}, this strategy achieves the best overall performance, especially on NC and TTC, demonstrating that multi-scale and iterative geometry aggregation is important for safe and reliable planning.

\begin{table*}[t]
\centering
\label{tab:planning_ablation}
\renewcommand{\arraystretch}{1.1}
\setlength{\tabcolsep}{4pt}

\begin{tabular}{c|c|cccccc}
\toprule

\textbf{Geo. Lay. Num.} 
& \textbf{Iter. Opt. Num.} 
& \textbf{NC$\uparrow$} 
& \textbf{DAC$\uparrow$} 
& \textbf{TTC$\uparrow$} 
& \textbf{Comf.$\uparrow$} 
& \textbf{EP$\uparrow$} 
& \textbf{PDMS$\uparrow$} \\

\midrule

24 & 1 & 98.5 & 95.7 & 95.1 & 99.7 & 81.5 & 87.6 \\
1 & 4 & 98.6 & 95.5 & 95.2 & 99.8 & \textbf{82.9} & 88.2 \\
\rowcolor{gray!15} 
4 & 4 & \textbf{98.9} & \textbf{97.2}  & \textbf{95.7} & \textbf{99.9} & 82.6 & \textbf{89.3} \\
\bottomrule
\end{tabular}
\caption{Quantitative comparison of three geometry aggregation strategies. \OURS employs a geometry-oriented planner that progressively aggregates multi-scale geometric features through iterative optimization. “Geo. Lay. Num.” denotes the number of geometry token layers used in the planner, while “Iter. Opt. Num.” denotes the number of iterative trajectory optimization steps.}\label{tab:geo_plan_interaction}
\end{table*}

\subsubsection{Geometry Evaluation of EgoStreamVGGT}

\begin{table*}[!t]
\centering
\renewcommand{\arraystretch}{1.1}
\setlength{\tabcolsep}{4pt}
\resizebox{0.95\textwidth}{!}{
\begin{tabular}{l l|cc|cc|cc}
\toprule
& & \multicolumn{2}{c|}{\textbf{OpenScene}} 
& \multicolumn{2}{c|}{\textbf{nuScenes}} 
& \multicolumn{2}{c}{\textbf{KITTI}} \\

\textbf{Method} & \textbf{Type} 
& \textbf{Abs Rel$\downarrow$} & $\boldsymbol{\delta < 1.25 \uparrow}$ 
& \textbf{Abs Rel$\downarrow$} & $\boldsymbol{\delta < 1.25 \uparrow}$ 
& \textbf{Abs Rel$\downarrow$} & $\boldsymbol{\delta < 1.25 \uparrow}$ \\
\midrule

StreamVGGT & Streaming & 0.236 & 65.6 & 0.265 & 58.2 & 0.173 & 72.2 
 \\
EgoStreamVGGT & Streaming & \textbf{0.141} & \textbf{86.5} & \textbf{0.117} & \textbf{88.5} & \textbf{0.077} & \textbf{95.5} \\
\bottomrule
\end{tabular}
}
\caption{Video depth estimation. Our finetuned EgoStreamVGGT achieves consistent improvements across OpenScene, nuScenes, and KITTI datasets.}\label{tab:depth_eval_supp}
\vspace{-3mm}
\end{table*}
\begin{table*}[t]
\centering

\renewcommand{\arraystretch}{1.1}
\setlength{\tabcolsep}{4pt}
\resizebox{0.95\textwidth}{!}{
\begin{tabular}{l l|ccc|ccc}
\toprule
& & \multicolumn{3}{c|}{\textbf{nuScenes}} 
& \multicolumn{3}{c}{\textbf{OpenScene}} \\

\textbf{Method} & \textbf{Type} 
& \textbf{ATE$\downarrow$} 
& \textbf{RPE trans$\downarrow$} 
& \textbf{RPE rot$\downarrow$} 
& \textbf{ATE$\downarrow$} 
& \textbf{RPE trans$\downarrow$} 
& \textbf{RPE rot$\downarrow$} \\

\midrule

StreamVGGT & Streaming 
& 14.79 & 1.77 & 0.47 & 8.66 & 1.00 & 1.53 \\
EgoStreamVGGT & Streaming & 5.78 & 0.63 & 1.31 & 4.07 & 0.39 & 0.92 \\
\bottomrule
\end{tabular}
}
\caption{Camera Pose Estimation Evaluation on nuScenes and OpenScene datasets.}
\label{tab:pose_estimation}
\end{table*}
To evaluate the effectiveness of the proposed ego-aligned geometry representation, we compare EgoStreamVGGT with the original StreamVGGT on both video depth estimation and camera pose estimation. The results are reported in Tab. \ref{tab:depth_eval_supp} and Tab. \ref{tab:pose_estimation}.

As shown in Tab. \ref{tab:depth_eval_supp}, EgoStreamVGGT consistently improves depth estimation performance across OpenScene, nuScenes, and KITTI. Compared with StreamVGGT, our model achieves lower absolute relative error and higher threshold accuracy on all three datasets. These results demonstrate that adapting the geometry backbone to the ego-centric driving setting effectively improves its ability to recover road-scene geometry. This improvement is especially important for autonomous driving, where accurate depth estimation helps the planner better understand drivable space, obstacle locations, and the spatial layout of surrounding agents. The consistent gains across different datasets also indicate that EgoStreamVGGT does not overfit to a single domain, but learns more robust geometry representations for diverse driving scenarios.

Tab. \ref{tab:pose_estimation} further evaluates camera pose estimation. EgoStreamVGGT significantly reduces trajectory-level and translational pose errors on both nuScenes and OpenScene, showing that the ego-aligned finetuning improves temporal geometry consistency in streaming video reconstruction. More accurate ego-motion estimation is crucial for GeoWorldAD because geometry tokens from different frames must be aligned in a consistent coordinate system before they can be used for planning. By reducing pose drift and frame-to-frame misalignment, EgoStreamVGGT provides more stable geometry features for downstream trajectory optimization. 

Overall, the geometry evaluation verifies that EgoStreamVGGT provides a stronger foundation for GeoWorldAD than the original StreamVGGT. Through ego-centric adaptation, the model obtains more accurate depth prediction and more consistent camera motion estimation, which together improve the quality of the geometry tokens used by the planner. These results support our design choice of building GeoWorldAD upon EgoStreamVGGT, as reliable geometry perception is essential for collision-aware and temporally consistent autonomous driving planning. Additional 4D reconstruction visualizations are provided in Sec. \ref{sec:vis_results}.

\clearpage
\subsection{Visualization Results}\label{sec:vis_results}

\subsubsection{4D Reconstruction Visualization}
\begin{figure*}[ht]
\centering
\includegraphics[width=1.00\textwidth]{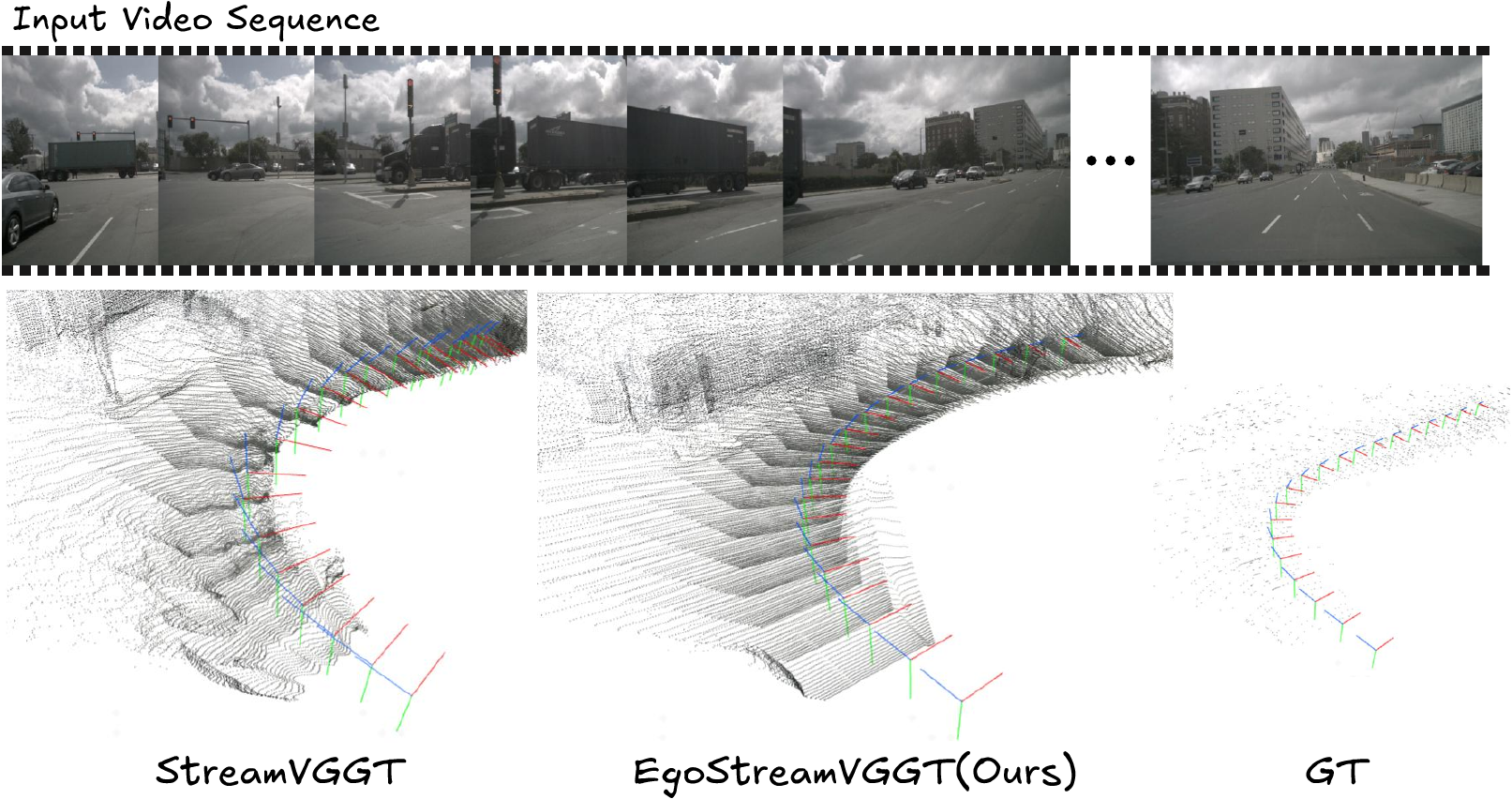}
\caption{Visual comparison of StreamVGGT and our EgoStreamVGGT for 4D reconstruction.}
\label{fig:recon_vis_0}
\end{figure*}

\begin{figure*}[ht]
\centering
\includegraphics[width=1.00\textwidth]{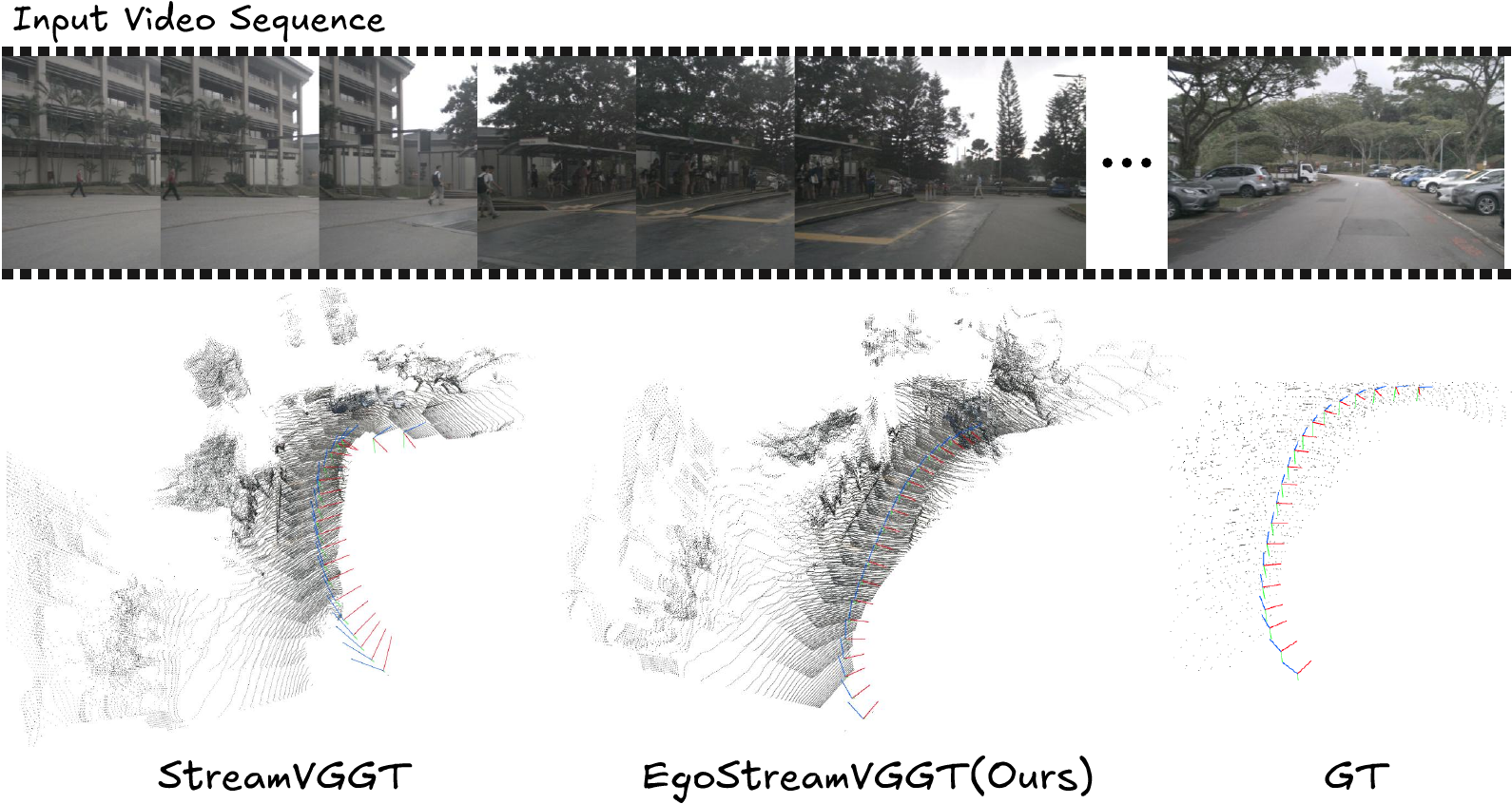}
\caption{Visualized comparison of StreamVGGT and our EgoStreamVGGT for 4D reconstruction.}
\label{fig:recon_vis_1}
\end{figure*}

\begin{figure*}[t]
\centering
\includegraphics[width=1.00\textwidth]{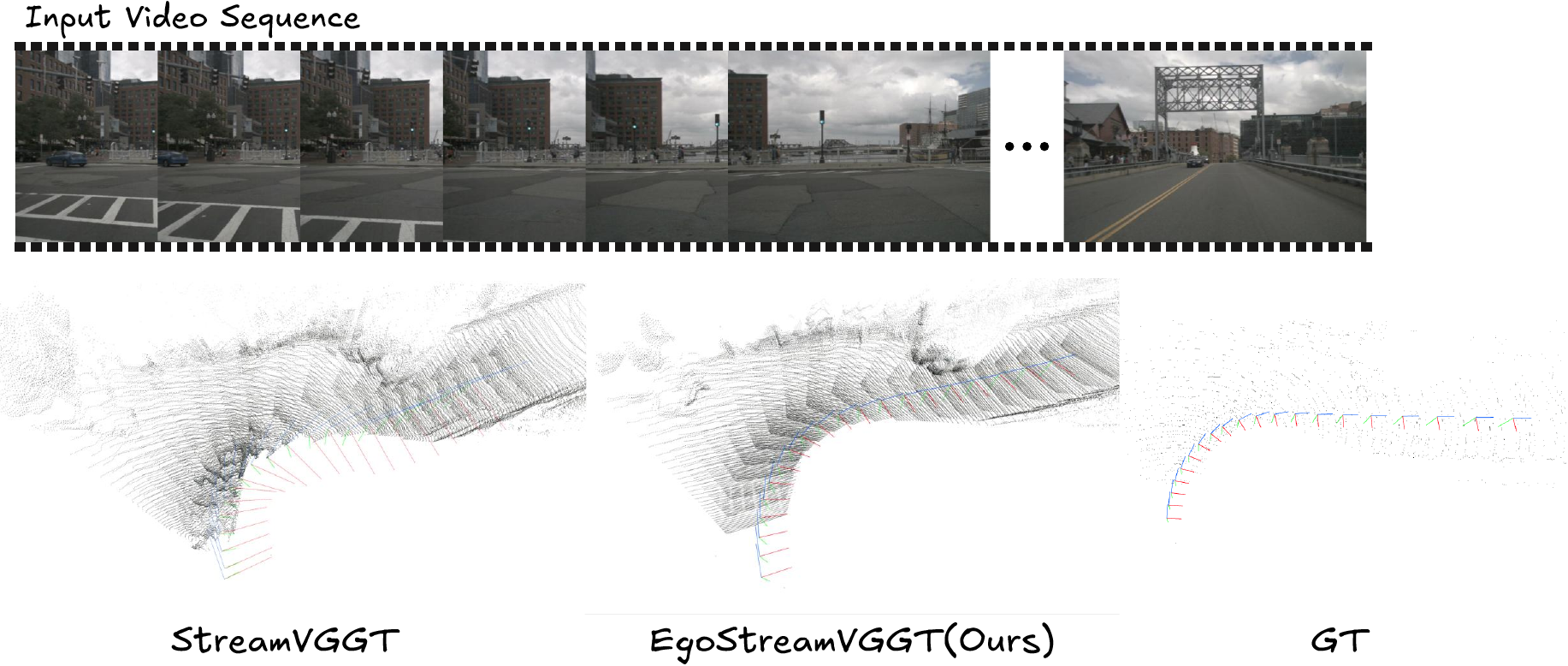}
\caption{Visualized comparison of StreamVGGT and our EgoStreamVGGT for 4D reconstruction.}
\vspace{-8cm}
\label{fig:recon_vis_2}
\end{figure*}

\begin{figure*}[t]
\centering
\includegraphics[width=1.00\textwidth]{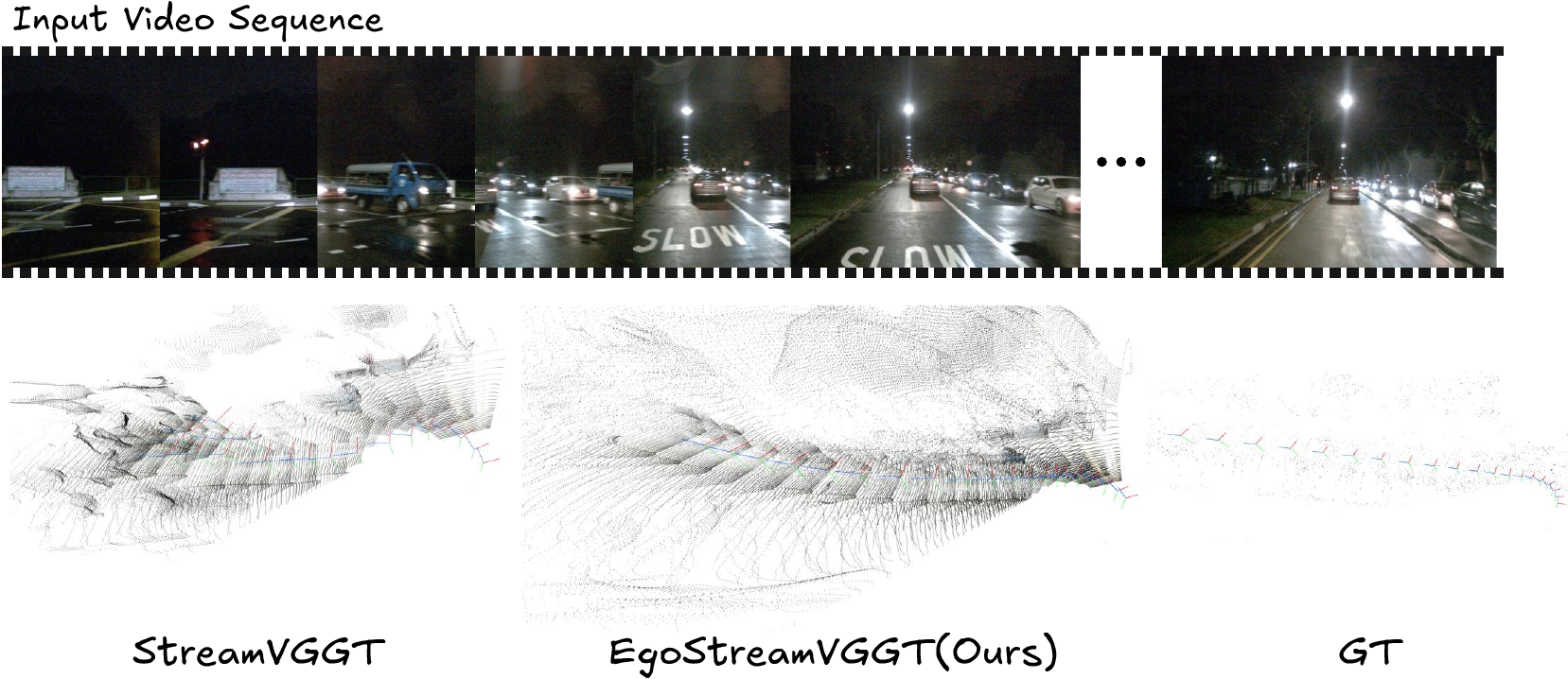}
\caption{Visualized comparison of StreamVGGT and our EgoStreamVGGT for 4D reconstruction.}
\label{fig:recon_vis_3}
\vspace{-2mm}
\end{figure*}

\clearpage
\subsubsection{Visualization of Future Depth Prediction}
\begin{figure*}[ht]
\centering
\includegraphics[width=0.95\textwidth]{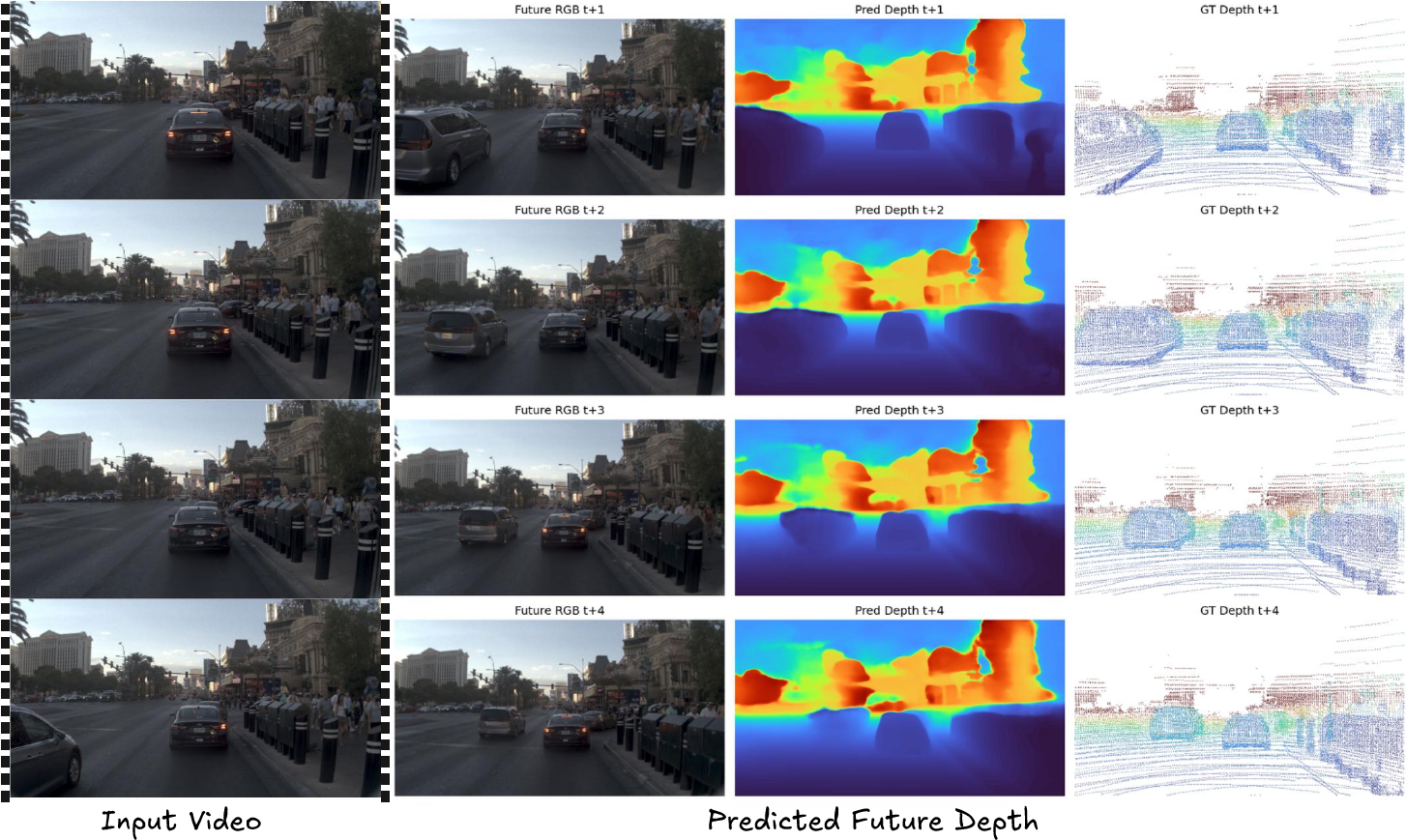}
\caption{Visualization of future depth prediction.}
\label{fig:supp_future_0}
\vspace{-2mm}
\end{figure*}

\begin{figure*}[ht]
\centering
\includegraphics[width=0.95\textwidth]{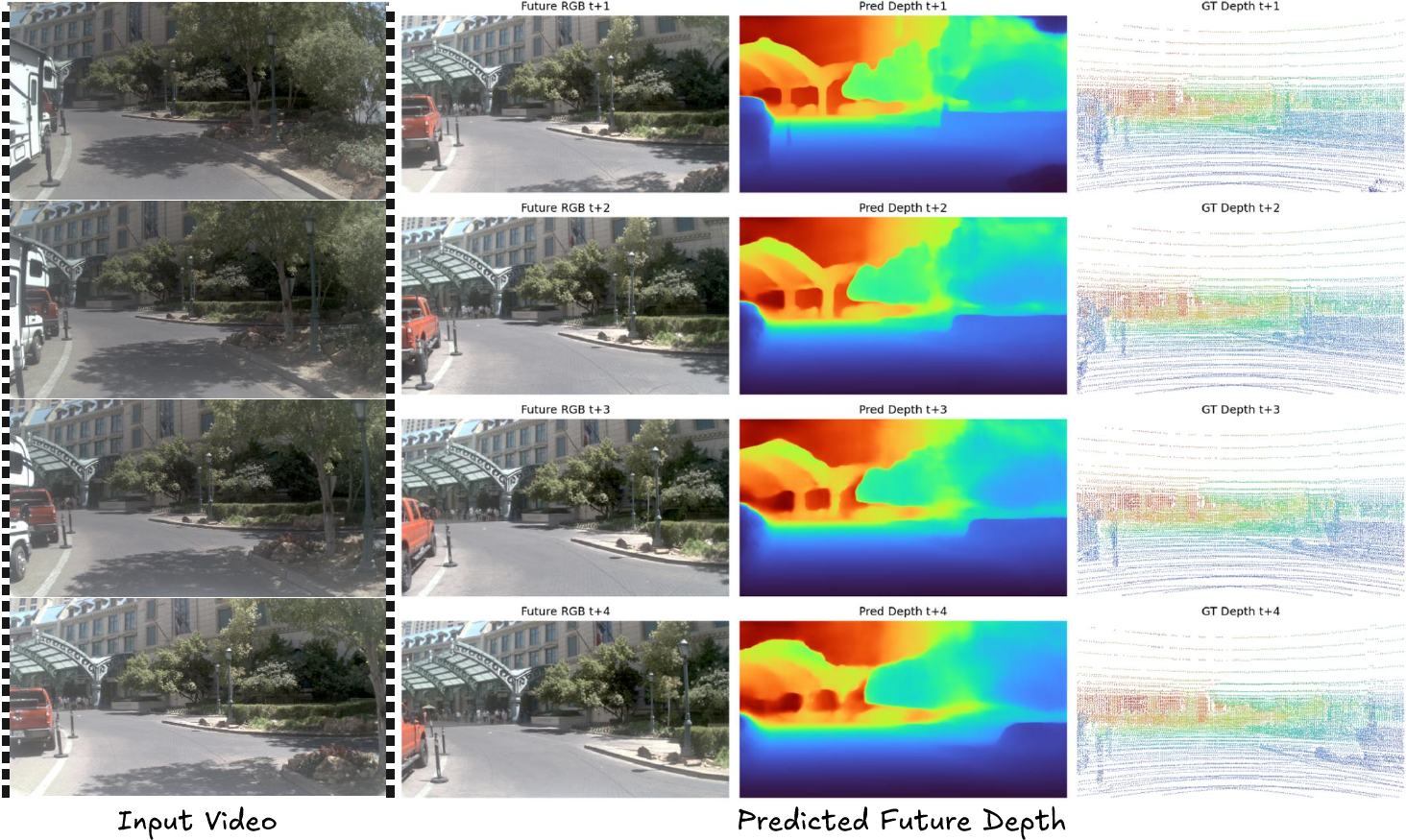}
\caption{Visualization of future depth prediction.}
\label{fig:supp_future_1}
\vspace{-2mm}
\end{figure*}

\end{document}